\pdfoutput=1
\documentclass[11pt]{article}

\newcommand{\jx}[1]{\color{black}{#1}}
\usepackage[preprint]{acl}

\usepackage{times}
\usepackage{latexsym}
\usepackage{pifont}% http://ctan.org/pkg/pifont
\newcommand{\cmark}{\ding{51}}%
\newcommand{\xmark}{\ding{55}}%
\usepackage{multirow}
\usepackage{amsmath}
\usepackage{bm}
\usepackage{subfigure}
\usepackage{amsfonts}
\usepackage[T1]{fontenc}
\usepackage[utf8]{inputenc}

\usepackage{microtype}

\usepackage{inconsolata}

\usepackage{graphicx}

\usepackage{booktabs}

\usepackage{enumitem}
\usepackage[most]{tcolorbox}

\tcbset{aibox/.style={
    breakable,
    break at=\maxdimen,
    fontlower=\scriptsize, 
}}
\newtcolorbox{AIbox}[2][]{aibox,title={#2},#1}

\title{Intention Knowledge Graph Construction for User Intention Relation Modeling}

\author{Jiaxin Bai\thanks{~~Equal contribution.}$^1$, Zhaobo Wang\footnotemark[1]$^2$, Junfei Cheng$^1$, Dan Yu$^1$, Zerui Huang$^1$, Weiqi Wang$^1$, \\\textbf{Xin Liu$^3$, Chen Luo$^3$, Yanming Zhu$^2$, Bo Li$^1$, and Yangqiu Song$^1$} \\
  $^1$CSE, Hong Kong University of Science and Technology \\
  $^2$CSE, Shanghai Jiaotong University \\
  $^3$ Amazon.com \\
  \texttt{\{jbai, jchengao, dyuao, zhuangeu, wwangbw, bli, yqsong\}@cse.ust.hk}, \\
  \texttt{ \{w19990112, yzhu \}@sjtu.edu.cn }, 
  \texttt{ \{xliucr, cheluo\}@amazon.com }
}

\begin{document}
\maketitle
\begin{abstract}
Understanding user intentions is challenging for online platforms. Recent work on intention knowledge graphs addresses this, but often lacks focus on connecting intentions, which is crucial for modeling user behavior and predicting future actions. This paper introduces a framework to automatically generate an intention knowledge graph, capturing connections between user intentions. Using the Amazon m2 dataset, we construct an intention graph with 351 million edges, demonstrating high plausibility and acceptance. Our model effectively predicts new session intentions and enhances product recommendations, outperforming previous state-of-the-art methods and showcasing the approach's practical utility.\footnote{ https://github.com/HKUST-KnowComp/RelationalIntentionGraph}
\end{abstract}

\begin{figure*}
    \begin{center}
\includegraphics[width=\linewidth]{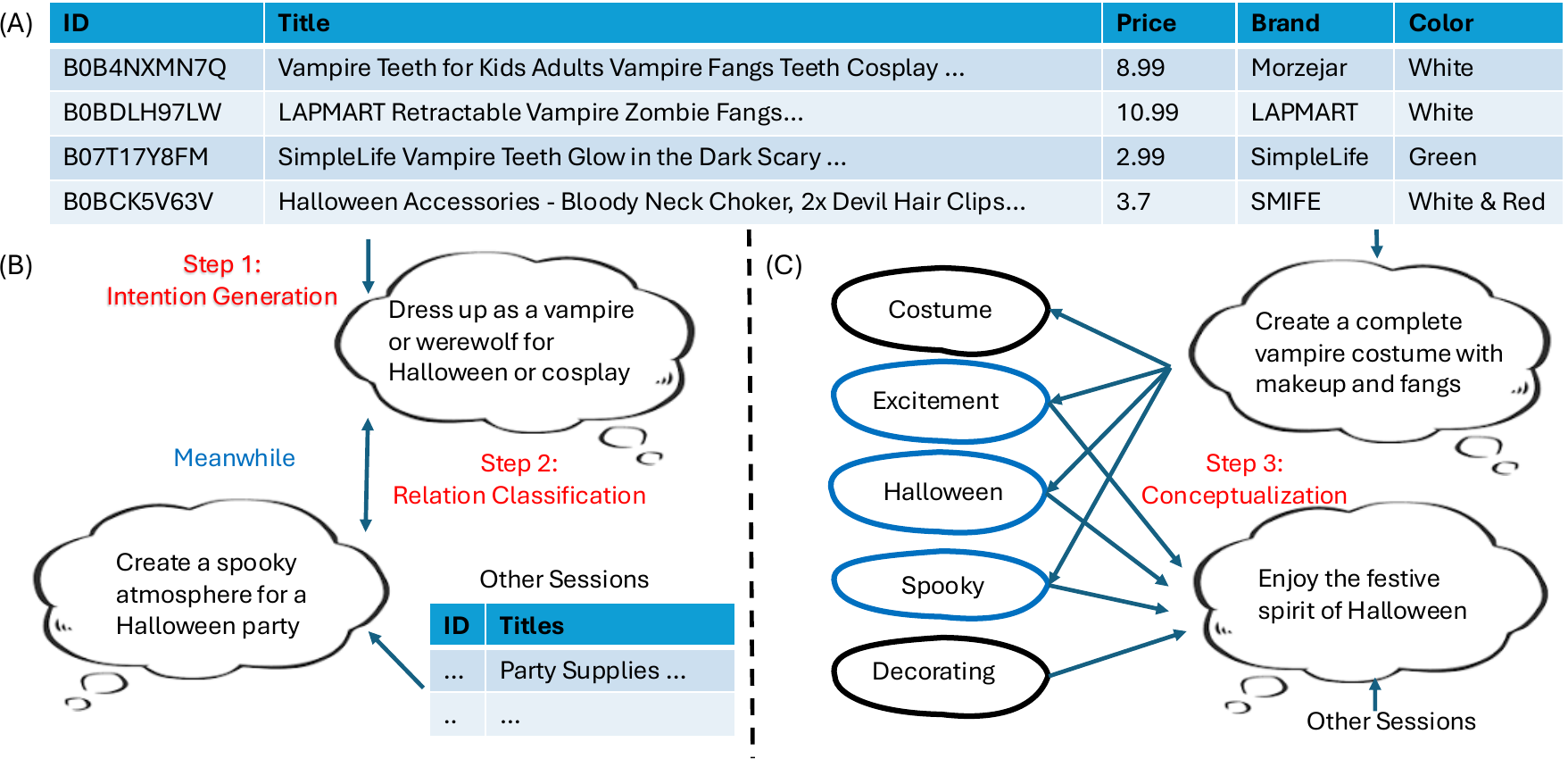}
\end{center}
\vspace{-0.2cm}
\caption{
The structure of our knowledge graph. Part (A) shows an example of user behaviors within a session. Our knowledge graph emphasizes building relations between different intentions.
In Part (B), we establish commonsense relations, highlighting the temporality and causality of intentions. Part (C) focuses on using conceptualization to connect various intentions.
}
\label{fig:teaser}
\vspace{-0.2cm}
\end{figure*}

\section{Introduction}
% Understanding user intentions behind behaviors poses a significant challenge for online platforms. Recent research has introduced the concept of intention knowledge graphs \cite{DBLP:conf/acl/YuWLBSLG0Y23, 10.1145/3626246.3653398} to address this issue. These graphs connect various user behaviors, such as co-purchasing \cite{DBLP:conf/acl/YuWLBSLG0Y23} and queries \cite{10.1145/3626246.3653398}, to their corresponding intentions expressed in natural language. intention knowledge graphs have proven valuable in applications like product recommendation \cite{DBLP:conf/acl/YuWLBSLG0Y23} and search relevance \cite{10.1145/3626246.3653398}.

Understanding user intentions is a fundamental challenge for online platforms seeking to optimize user experience. Recent advances in intention knowledge graphs have made significant strides in connecting user behaviors to their underlying intentions. These graphs link various behaviors, such as co-purchasing patterns \cite{DBLP:conf/acl/YuWLBSLG0Y23} and search queries \cite{10.1145/3626246.3653398}, to intentions expressed in natural language, proving valuable for applications ranging from product recommendation to search relevance optimization.

However, a critical limitation of existing approaches lies in their inability to model the relationships between different intentions. Consider a user preparing for Halloween—they may have multiple interconnected intentions: dressing as a vampire, creating spooky decorations, and hosting a party. While their browsing history (Figure \ref{fig:teaser} (A)) implicitly suggests these intentions, existing knowledge graphs fail to capture how these intentions relate to each other through temporal, causal, and conceptual links.

This gap aligns with Bengio's distinction between System I and System II reasoning, where the latter emphasizes deliberate, logical, and sequential thinking. Research demonstrates that online shopping behaviors frequently reflect System II reasoning \cite{DBLP:conf/sigecom/KleinbergMR22}, with users making conscious decisions toward long-term goals rather than engaging in unconscious browsing. By explicitly modeling intention relationships, we can better capture the deliberate nature of user behavior and create more accurate predictive models.

% Users often have specific intentions, such as preparing for a Halloween party. They may want to dress up as a vampire or werewolf and create a spooky atmosphere with decorations. However, these intentions are often only implicitly suggested by actions, such as the items shown in Figure \ref{fig:teaser} (A). Previous attempts at constructing intention knowledge graphs have focused on using large language models to generate explanations for behaviors \cite{DBLP:conf/acl/YuWLBSLG0Y23}, utilizing sources like session history and query keywords \cite{10.1145/3626246.3653398}. For instance, analyzing the browsing history of items can reveal the intention to dress up for Halloween.

% However, previous methods have not explored how intentions interrelate. Explicitly modeling these connections aligns with the shift towards System II \footnote{\url{http://www.iro.umontreal.ca/~bengioy/AAAI-9feb2020.pdf}} reasoning, emphasizing logical and sequential reasoning. Research has shown the importance of this reasoning in online behaviors \cite{DBLP:conf/sigecom/KleinbergMR22}, where users focus on long-term rewards rather than unconscious browsing. Therefore, we aim to model explicit intention relations.

Our research extends beyond merely understanding users' initial intentions to predicting subsequent ones, for instance, inferring that someone who intends to buy a desk may soon need an office chair. This predictive capability has profound implications for user behavior modeling, session understanding, and recommendation systems. However, constructing these intention-to-intention relationships remains unexplored in current research.

% Our goal extends beyond understanding initial intentions to predict subsequent ones. For example, intending to buy a desk might lead to buying an office chair. This inference can improve user behavior modeling and recommendations. However, building these connections is unexplored.

We identify two key mechanisms for modeling these relationships. First, commonsense knowledge is essential for understanding how intentions connect—planning a Halloween party naturally entails both costume selection and decoration preparation. We propose leveraging inferential commonsense relations to describe temporal sequences (before/after) and causal connections (because/as a result) between intentions. After identifying intentions from user sessions, we employ a classifier to determine these inferential connections, building a more comprehensive model of user intention dynamics. Figure \ref{fig:teaser} (B) illustrates this approach, showing the plausible co-occurrence relationship between costume and decoration intentions.

Second, we incorporate abstract conceptualization to improve generalization beyond observed intention pairs. Recent work demonstrates that conceptualization and instantiation enhance commonsense reasoning \cite{DBLP:conf/acl/WangFXBSC23, DBLP:conf/emnlp/WangF0XLSB23}. We integrate this insight by using models to abstract intentions into broader concepts, establishing connections between semantically related intentions. For example, conceptualizing a "vampire costume" intention involves higher-level concepts like "costumes" and "Halloween," linking it to other costume-related intentions (Figure \ref{fig:teaser} (C)).

% Commonsense knowledge is crucial for modeling intention relationships. For example, planning a Halloween party requires understanding that costumes and decorations are likely intentions. We propose using commonsense relations to describe temporal and causal intention connections. After identifying intentions from user sessions, we use a classifier to determine inferential connections, enhancing our understanding of intention relationships. For instance, Figure \ref{fig:teaser} (B) demonstrates the plausible co-occurrence relationship between dressing up and decorating intentions.

% Using inferential commonsense relations can be challenging to generalize to unseen situations. Concepts like conceptualization and instantiation help generalize commonsense reasoning \cite{DBLP:conf/acl/WangFXBSC23, DBLP:conf/emnlp/WangF0XLSB23, DBLP:journals/corr/abs-2311-09174, DBLP:journals/corr/abs-2401-07286}. We incorporate this into our knowledge graph, using models to conceptualize intentions into broader concepts, connecting related intentions. For example, conceptualizing the intention to create a vampire costume can involve concepts like costumes and Halloween, linking it to similar intentions.

To address these challenges, we introduce the \textbf{Intention Generation, Conceptualization, and Relation Classification (IGC-RC)} framework for constructing a comprehensive commonsense knowledge graph of user intentions. Our approach follows three key steps:
\textbf{(1) Intention Generation:} We leverage large language models to generate plausible intentions from user session data, capturing the underlying goals driving observable behaviors.
 \textbf{(2) Conceptualization:} We abstract these intentions into higher-level concepts, facilitating connections between semantically related intentions.
\textbf{(3) Relation Classification:} We generate and verify commonsense statements describing relationships between intentions, creating a structured graph of interconnected user goals.

% We introduce the {Intention Generation, Conceptualization, and Relation Classification (IGC-RC)} framework for constructing a commonsense knowledge graph from user behaviors to address the challenges. The framework involves three general steps:
% (1) \textbf{Intention Generation:} Using large language models, we generate intentions from user sessions.
% (2)\textbf{ Conceptualization:} We conceptualize the generated intentions using conceptualization models.
% (3)\textbf{ Relation Classification:} We generate commonsense statements from intention relations and verify their plausibility.

We apply the IGC-RC framework to the Amazon M2 session-based recommendation dataset, constructing a Relational Intention Graph (RIG) with 351 million edges. This graph captures rich intention-level relationships and demonstrates high plausibility in human evaluations. Our extensive experiments show that RIG enables accurate prediction of intentions in new user sessions and significantly enhances the performance of session-based recommendation models, outperforming previous state-of-the-art approaches.

Our contributions include:
\begin{itemize}
    \item We pioneer the modeling of connections between users' deliberate mental processes through an intention knowledge graph that captures temporal, causal, and conceptual relationships between intentions.
    \item We develop the IGC-RC framework, a novel methodology that integrates user behavior data with large language models to automatically construct rich, multi-faceted intention knowledge graphs.
    \item We construct RIG, a large-scale, high-quality intention knowledge graph from the Amazon M2 dataset, demonstrating superior performance in session understanding and recommendation tasks.
\end{itemize}

By bridging the gap between discrete user behaviors and their interconnected intentions, our work represents a significant advancement in user modeling for e-commerce and recommendation systems, with broad implications for creating more responsive and intuitive online platforms.

\begin{table*}[th]
\resizebox{\linewidth}{!}{\begin{tabular}{@{}lccccccc@{}}
\toprule
KG & \# Nodes & \# Edges & \# Rels & Sources & Node Type & Intention Relations & User Behavior \\ \midrule
ConceptNet & 8M & 21M & 36 & Croudsource & concept & \cmark & \xmark \\
ATOMIC & 300K & 870K & 9 & Croudsource & situation, event & \cmark & \xmark \\
AliCoCo & 163K & 813K & 91 & Extraction & concept & \xmark & search logs \\
AliCG & 5M & 13.5M & 1 & Extraction & concept, entity & \xmark & search logs \\
FolkScope & 1.2M & 12M & 19 & LLM Generation & product, intention & \xmark & co-buy \\
COSMO & 6.3M & 29M & 15 & LLM Generation & product, query, intention & \xmark & co-buy search-buy \\
RIG (Ours) & 4.2M & 351M & 6 & LLM Generation & product, session, intention, concept & \cmark & session item history \\ \bottomrule
\end{tabular}
}
\vspace{-0.3cm}
\caption{This table shows the details and differences between different commonsense knowledge graphs. { Our graph contains six distinct types of edges, including three types of intention-to-intention relationships: asynchronous (before/after), synchronous (at the same time), and causality (because/as a result) among intentions, and item-to-session, session-to-intention, and intention-to-concept connections, summing up to six edge types.}}
\vspace{-0.3cm}
\end{table*}

\begin{table*}[t]
% \resizebox{\linewidth}{!}{
\centering
\small
\begin{tabular}{@{}cccccccc@{}}
\toprule
\# Sessions & \# Concepts & \# Intentions & \# Ses.-Int. & \# Int.-Con. & \# Int.-Int. & \# Nodes & \# Edges \\ \midrule
1,176,296 & 110,741 & 2,956,195 & 5,115,587 & 5,115,212 & 341,649,216 & 4,243,232 & 351,880,015 \\ \bottomrule
\end{tabular}
% }
\vspace{-0.3cm}
\caption{The overall statistics of our constructed knowledge graph RIG, including sessions, intentions, and concepts. Our knowledge graph includes 351 million edges.}
\vspace{-0.4cm}
\end{table*}

\section{Related Work}

E-commerce platforms increasingly rely on knowledge graphs (KGs) for personalized experiences. Notable examples include Amazon Product Graph \citep{DBLP:conf/kdd/ZalmoutZLLD21}, which aligns with Freebase \citep{DBLP:conf/sigmod/BollackerEPST08}, and Alibaba's ecosystems: AliCG \citep{DBLP:conf/kdd/ZhangJD0YCTHWHC21}, AliCoCo \citep{DBLP:conf/sigmod/LuoLYBCWLYZ20}, and AliMeKG \citep{DBLP:conf/cikm/LiCXQJZC20}. While these KGs effectively represent item properties, they typically miss user intentions.

Recent work by FolkScope \cite{DBLP:conf/acl/YuWLBSLG0Y23} and COSMO \cite{10.1145/3626246.3653398} incorporates user intentions through large language models, but lacks inter-intention relationships crucial for modeling goal-oriented behavior. Commonsense knowledge graphs like ConceptNet \citep{DBLP:conf/aaai/SpeerCH17}, ATOMIC \citep{DBLP:conf/aaai/SapBABLRRSC19,DBLP:conf/aaai/HwangBBDSBC21}, Discos \citep{DBLP:conf/www/FangZWSH21}, and WebChild \citep{DBLP:conf/acl/TandonMW17} structure general knowledge but don't address e-commerce-specific intentions.

Session-based recommendation has evolved from sequence modeling approaches using RNNs and CNNs \citep{hidasi2015session,li2017neural,DBLP:conf/kdd/LiuZMZ18,DBLP:conf/wsdm/TangW18} to Graph Neural Networks \citep{DBLP:conf/sigir/LiPLSLXYCZZ21,DBLP:conf/cikm/GuoZL0ZK22,DBLP:conf/icml/HuangWLH22}. \citet{DBLP:conf/aaai/WuT0WXT19} pioneered GNNs for modeling session transitions, with subsequent improvements incorporating contextual information \citep{DBLP:conf/cikm/PanCCCR20,DBLP:conf/aaai/0013YYWC021}. However, these methods often simplify user intentions, whereas our approach models them explicitly within their relational context.

Meanwhile, knowledge graphs have also demonstrate its potential in solving various types of complex logical questions \cite{DBLP:conf/naacl/BaiWZS22, bai2023sequential, DBLP:conf/kdd/BaiLLYYS23, DBLP:conf/nips/BaiLW0S23, DBLP:conf/acl/BaiWZGLS24, DBLP:conf/acl/ZhengW0B0DS025} in e-commerce domain \cite{DBLP:conf/kdd/Bai0LYS24}. Automatic knowledge graph construction with intentions can also provide resource to the construction of neural graph databases \cite{DBLP:conf/log/BestaISODPCH22, DBLP:journals/corr/abs-2303-14617, DBLP:conf/kdd/Hu0BWS24, DBLP:journals/debu/Bai0Z0FHDCZTGXL25, DBLP:journals/tmlr/HuJLWBMSFL25}. 

\begin{table*}[t]
\resizebox{\linewidth}{!}{
\begin{tabular}{@{}ll@{}}
\toprule
Intention & Concepts \\ \midrule
Purchase a construction dump truck toy for a 2-year-old boy or girl. & playtime, construction, gift \\
Enjoy multiplayer gameplay with friends and family & socializing, competition, and fun. \\ 
Personalize their drawstring bags. & personalization, gift, accessorizing \\
Perform sanding and grinding tasks on large surfaces using an orbital sander. & smoothing, surface prep, precision \\
Improve their precision cutting skills & precision, sharpness, craftsmanship \\ \bottomrule
\end{tabular}}

\caption{This table maps user intentions to relevant concepts. Each intention is analyzed to highlight the core concepts, showcasing how these insights can inform personalized recommendation systems.}
\label{tab:conceptulization_example}

\end{table*}

% Recent methods have explored using session history to reflect user intentions and enhance recommendation systems. Sequence modeling approaches, including Recurrent Neural Networks (RNNs) and Convolutional Neural Networks (CNNs), have been employed to model session data \citep{hidasi2015session,li2017neural, DBLP:conf/kdd/LiuZMZ18, DBLP:conf/wsdm/TangW18}.
% Advancements in session-based recommendation systems have focused on Graph Neural Networks (GNNs) to capture better session transitions \citep{DBLP:conf/sigir/LiPLSLXYCZZ21,DBLP:conf/cikm/GuoZL0ZK22,DBLP:conf/icml/HuangWLH22}. \citet{DBLP:conf/aaai/WuT0WXT19} first utilized GNNs to capture complex transitions using graph structures. Subsequent research incorporated position, target information, global context, and highway networks to enhance performance \citep{DBLP:conf/cikm/PanCCCR20,DBLP:conf/aaai/0013YYWC021}.
% However, existing methods often simplify user intentions by focusing on item features or statistical metrics, resulting in an inadequate understanding of user preferences. Our approach aims to address this by explicitly modeling user intentions.

\section{IGC-RC Framework}

We introduce the Intention Generation, Conceptualization, and Relation Classification (IGC-RC) framework for constructing comprehensive intention knowledge graphs from user behavior data. Our approach integrates large language models with structured knowledge representation techniques to build a rich understanding of user intentions and their relationships.
\label{sec:methodology}
\subsection{Intention Generation}

% We utilize the Amazon M2 dataset \cite{DBLP:conf/nips/0009MLJ0WHLW0LC23}, focusing on the English subset with 1.2 million sessions, to generate user intentions. The session history, comprised of sequences of user-browsed products, is processed to extract attributes such as titles, descriptions, models, sizes, and colors. These details are compiled into a JSON file, serving as the input for our intention generation module.

% Using GPT-3.5, we generate concise, informative, and diverse user intentions, totaling 4.3 million from the dataset. Our approach surpasses FolkScope by using an advanced language model and eliminating ConceptNet relation constraints, fostering a more diverse output. See Figure \ref{fig:generation_prompts} in the appendix for our prompt designs. 
% We also include an evaluation on intention diversity in the Appendix Figure \ref{tab:ngram_diversity}.

We leverage the Amazon M2 dataset \cite{DBLP:conf/nips/0009MLJ0WHLW0LC23}, specifically its English subset containing 1.2 million sessions. For each session, we extract product attributes including titles, descriptions, specifications, and technical details to create a comprehensive representation of user browsing history.

Using GPT-3.5, we generate 4.3 million diverse user intentions from these session histories. Our approach improves upon previous work by employing a more capable language model and removing restrictive relation constraints, resulting in more natural and contextually appropriate intentions. The prompting strategy (detailed in Appendix Figure \ref{fig:generation_prompts}) focuses on extracting concise verb phrases that capture user goals underlying their browsing behavior. Quantitative analysis in Appendix Figure \ref{tab:ngram_diversity} confirms significantly higher n-gram diversity compared to Folkscope because of using a better generation model and removal of conceptnet relation constraints.

\begin{table*}[t]
\resizebox{\linewidth}{!}{
\begin{tabular}{@{}l|ccc|ccc|ccc|ccc|ccc|ccc@{}}
\toprule
 & \multicolumn{3}{c|}{Precedence} & \multicolumn{3}{c|}{Succession} & \multicolumn{3}{c|}{Simultaneous} & \multicolumn{3}{c|}{Cause} & \multicolumn{3}{c|}{Result} & \multicolumn{3}{c}{Overall} \\
 & Prec. & Rec. & F1 & Prec. & Rec. & F1 & Prec. & Rec. & F1 & Prec. & Rec. & F1 & Prec. & Rec. & F1 & Prec. & Rec. & F1 \\ \midrule
Vera & 0.73 & 0.52 & 0.61 & 0.78 & 0.66 & 0.72 & 0.75 & 0.83 & 0.79 & 0.75 & 0.73 & 0.74 & 0.76 & 0.92 & 0.83 & 0.75 & 0.73 & 0.74 \\
+ Fine-tuning & \textbf{0.86} & \textbf{0.97} & \textbf{0.91} & \textbf{0.88} & \textbf{0.92} & \textbf{0.90} & \textbf{0.87} & \textbf{0.93} & \textbf{0.90} & \textbf{0.84} & \textbf{0.91} & \textbf{0.88} & \textbf{0.81} & \textbf{0.99} & \textbf{0.89} & \textbf{0.85} & \textbf{0.94} & \textbf{0.89} \\ \bottomrule
\end{tabular}}
\vspace{-0.3cm}
\caption{The performance of the VERA classifier on the annotated relation classification task. }
\vspace{-0.3cm}
\label{tab:vera_finetune}
\end{table*}

\subsection{Intention Relation Classification}

Creating meaningful connections between intentions requires understanding their complex logical, temporal, and causal relationships. We develop a three-step approach to establish these connections: \textbf{(1) Template-based assertion generation:} We transform intention pairs into natural language assertions using templates that express five relationship types: precedence, succession, simultaneity, cause, and result (examples in Table \ref{tab:vera_examples}).
\textbf{(2) Plausibility estimation:} We employ the Vera model \citep{liu-etal-2023-vera}, fine-tuned on expert-annotated data, to assess the plausibility of these assertions. The annotation process followed the Penn Discourse Treebank 2.0 guidelines \cite{prasad-etal-2008-penn}, with two experts independently evaluating each sample (85\% initial agreement). (3) T\textbf{hreshold-based edge selection:} We retain only high-confidence relationships (Vera score > 0.9) to ensure graph quality. The fine-tuned model achieves 0.89 F1 score overall (Table \ref{tab:vera_finetune}), with particularly strong performance on temporal relationships.

This approach enables us to model diverse intention relationships including temporal sequence (before/after), co-occurrence (simultaneously), and causality (because/results in).

\subsection{Intention Conceptualization}
% A conceptualization provides an abstract, simplified view of a selected part of the world, encompassing objects, concepts, and relationships \cite{gruber1993translation,himanen2019data}. Intention conceptualization extracts abstract concepts from user intentions to represent them in a knowledge graph, aiding e-commerce recommendation systems by modeling user intention dynamics.

% Recent studies \cite{DBLP:conf/acl/WangFXBSC23,DBLP:conf/emnlp/WangF0XLSB23,DBLP:journals/corr/abs-2311-09174,DBLP:journals/corr/abs-2401-07286} primarily address entity and event conceptualization, leaving user intention conceptualization as an open challenge. Two main approaches exist: crowd-sourced and large language model (LLM) annotations \cite{DBLP:journals/corr/abs-2401-07286}. While crowd-sourcing offers high-quality results, it is costly and limited in coverage. Thus, we utilize a large language model for intention conceptualization.

% To ensure diverse, unambiguous conceptualizations for machine integration, we design specific prompts to guide the LLM. Using the \texttt{Meta-Llama-3-8B-Instruct} model, we generate conceptualizations for each intention. The prompts used are detailed in Figure~\ref{fig:conceptualization_prompts}, and examples of the generated conceptualizations are shown in Table~\ref{tab:conceptulization_example}.

To improve generalization and abstraction, we develop techniques to map specific intentions to broader concepts. Building on theoretical foundations from knowledge representation \cite{gruber1993translation,himanen2019data} and recent advances in conceptualization \cite{DBLP:conf/acl/WangFXBSC23,DBLP:journals/corr/abs-2401-07286}, we create a novel approach specifically for intentions.

We employ Meta-Llama-3-8B-Instruct with carefully designed prompts (Appendix Figure \ref{fig:conceptualization_prompts}) to generate concise, non-redundant concept sets for each intention. The prompt emphasizes three key qualities: (1) representativeness: concepts must accurately capture the intention's essence; (2) unambiguity: concepts should have clear, focused meanings; and (3) complementarity: concepts should cover different semantic aspects of the intention.

Table \ref{tab:conceptulization_example} demonstrates the effectiveness of this approach, showing how diverse intentions are mapped to concise concept sets that capture their essential characteristics. These concepts serve as connective tissue in our knowledge graph, enabling us to link semantically related intentions that might otherwise appear unrelated based on surface features.

By integrating these three components, intention generation, relation classification, and conceptualization, our IGC-RC framework creates a rich, structured representation of user intentions and their relationships.

\section{Intrinsic Evaluation}

In this section, we evaluate the quality of the generated knowledge graph using both crowdsourced human annotations and automatic evaluations.

\subsection{Human Evaluation}

\begin{table}[t]
\centering

\begin{tabular}{@{}lcc@{}}
\toprule
Knowledge Graph & Plausibility & Typicality \\ \midrule
FolkScope {\jx (Before Filter)} & 0.6116 & 0.4491 \\ \midrule
RIG (Ours) & \textbf{0.9552} & \textbf{0.6674} \\
+ ConceptNet Rels  & {\jx 0.9542} & {\jx 0.6430} \\
\bottomrule

\end{tabular}
\vspace{-0.3cm}
\caption{The intention generation quality comparison with FolkScope. {\jx The comparison is \textbf{only} used to show the quality of intention generation, rather than the overall quality of two knowledge graphs.}}
\vspace{-0.5cm}
\label{tab:intention_quality}
\end{table}
\paragraph{Intention Generation}
Annotators evaluate two aspects of the generated intentions: plausibility and typicality. \textbf{Plausibility} refers to the likelihood that an assertion is valid based on its properties, usages, and functions. \textbf{Typicality} measures how well an assertion reflects specific features influencing user behavior, such as informativeness and causality. For example, "they are used for Halloween parties" is more informative than "they are used for the same purpose."
{ We collected annotations for 3,000 session-intention pairs, each evaluated by three annotators. The inter-annotator agreement scores were 0.91 for plausibility and 0.74 for typicality.
We use \textit{the same} annotation guidelines and criteria from the FolkScope paper, and this can ensure the same standard of annotations on plausibility and typicality.  
}
As shown in Table \ref{tab:intention_quality}, our RIG surpasses FolkScope~\citep{DBLP:conf/acl/YuWLBSLG0Y23}  in plausibility and typically.
{ Moreover, as shown in Table \ref{tab:vera_examples}, the model successfully captures particular, long-tail intentions such as "Relieve discomfort and soothe itching caused by hemorrhoids" and "Purchase unscented baby wipes for sensitive skin." These examples illustrate the model's ability to understand and articulate context-specific user needs. } 
{\jx Meanwhile, adding ConceptNet relations as constraints is not useful for plausibility also negatively impacts the intention typicality.}

\paragraph{Intention Relation Classification}

\begin{table}[t]
\centering
\resizebox{0.9\linewidth}{!}{
\begin{tabular}{@{}lcc@{}}
\toprule
Knowledge Graph & Acceptance & Size \\ \midrule
Atomic2020 & \textbf{86.8} & 0.6M \\
Atomic10x & 78.5 & 6.5M \\
Atomic NOVA & - & 2.1M \\
RIG (ours) & \underline{81.2} & \textbf{341.6M} \\
\thickspace - Asynchronous Relation & 80.6 & 100.2M \\
\thickspace - Synchronous Relation & 82.8 & 112.6M \\
\thickspace - Causality Relation & 80.4 & 128.7M \\
\bottomrule
\end{tabular}}
\vspace{-0.2cm}
\caption{This table presents a comparative analysis of our proposed knowledge graph with existing commonsense knowledge graphs in terms of relation plausibility acceptance and scale of relation edges. As evaluated by human annotators, the acceptance rate represents the percentage of plausible relations in each knowledge graph. The size of each knowledge graph is measured in millions (M) of relation edges.}
\label{tab: size}
\vspace{-0.4cm}
\end{table}

Annotators rate the plausibility of predicted intention relations on a four-point scale: Plausible, Somewhat plausible, Not plausible, and Not applicable. The first two are deemed acceptable. 
{ For intention relation classification, three annotators independently evaluated 1,000 intention-intention discourse pairs, achieving an overall inter-annotator agreement of 0.69.}
As shown in Table \ref{tab: size}, our graph achieves an acceptance rate of 81.2\% with a significantly larger scale than previous graphs. Atomic2020 has a higher acceptance rate due to manual creation and annotation.

\paragraph{Intention Conceptualization}

In this task, annotators evaluate the correctness of conceptualized intentions derived from user sessions. We sample 1,000 intention-concept pairs and use Amazon Mechanical Turk for annotation. The conceptualization performance is 86.60\%, with an inter-annotator agreement of 77.19\%.

\subsection{Automatic Evaluation}

In this section, we systematically evaluate three key aspects:
(1) Intention Prediction;
(2) Conceptualization of New Intentions; and
(3) Item Recovery.
These experiments are termed ``automatic evaluation'' as they can uniformly be regarded as inductive knowledge graph completion tasks.

\paragraph{Intention Prediction}

\begin{table}[t]
\resizebox{\linewidth}{!}{
\begin{tabular}{@{}llllll@{}}
\toprule
 & MRR & Hit@1 & Hit@3 & Hit@10 & Inf. Time \\ \midrule
Llama3-8B & 0.4680 & 0.4062 & 0.4480 & 0.5879  & 4,102.92ms \\
Mistral-7B & \textbf{0.5544} & \textbf{0.4954} & 0.5425 & 0.6763  & 3,625.63ms \\
Flan-T5 & 0.1575& 0.0528 & 0.1295 & 0.3642  & 2,021.36ms \\ \midrule
RIG (ours) & 0.5377& 0.4483 & \textbf{0.5470}
 & \textbf{0.7260}  & \textbf{3.01ms} \\ \bottomrule
\end{tabular}}
\vspace{-0.2cm}
\caption{The performance on intention prediction.  }
\vspace{-0.4cm}
\label{tab:intention_prediction}
\end{table}

% This study evaluates the task of linking unseen sessions to an existing intention graph. The dataset, comprising all session-intention edges, was split into training, validation, and test sets (8:1:1) {\jx make it more clear, I do not really know how to make it more clear already. Maybe draw a figure.}, with each session linked to 2-4 positive intentions. For evaluation, we applied random negative sampling to generate 30 candidate intentions per session, ranking the positive intentions among them.

% We compared large language models ({Mistral-7B-Instruct-v0.1}, {Meta-Llama-3-8B-Instruct}, and {flan-t5-large}) to our RIG model. Baselines use perplexity scores derived via proper prompting to rank intentions, while RIG employs an embedding-based approach. Specifically, we generate embeddings for sessions and intentions using a sentence model and use SASRec as a session encoder to compute session representations.

{\jx This study evaluates the task of linking unseen sessions to an existing intention graph. The dataset, comprising all session-intention edges, was split into training, validation, and test sets (8:1:1) based on sessions, with each session linked to 2-4 positive intentions. For evaluation, we applied random negative sampling to generate 30 candidate intentions per session, ranking the positive intentions among them.
We compared large language models ({Mistral-7B-Instruct-v0.1}, {Meta-Llama-3-8B-Instruct}, and {flan-t5-large}) to our RIG model. These models are open-sourced and has the licenses that allow academic use. Baselines use perplexity scores derived via proper prompting to rank intentions, while RIG employs an embedding-based approach. Specifically, we generate embeddings for sessions and intentions using a sentence model and use SASRec as a session encoder to compute session representations.}
As shown in Table~\ref{tab:intention_prediction}, our RIG model achieves competitive accuracy (e.g., best Hit@10) while significantly outperforming LLMs in inference speed (3.01ms vs. 2,021–4,102ms), demonstrating its practicality for real-world applications requiring fast decision-making.

\paragraph{Conceptualization Prediction}

{\jx 
To evaluate the performance of conceptualization prediction, we constructed a dataset of intention-concept pairs derived from our intention knowledge graph. This dataset was split into training, validation, and test sets using an 8:1:1 ratio. The test set consisted of 147,801 intention-concept pairs, which were used to perform ranking tasks with various methods.
For the baseline models, we employed large language models (LLMs) to perform ranking in a generative manner. Specifically, we experimented with \texttt{Mistral-7B-Instruct-v0.3}, \texttt{Meta-Llama-3-8B-Instruct}, and \texttt{flan-t5-xl}. For each intention, a candidate pool containing both true and false concepts was constructed, and the concepts were ranked based on their generation order by the LLMs.
Similar to the intention prediction task, we applied negative sampling for each intention to rank positive concepts among a pool of 500 candidates. Our proposed method relied on an embedding-based approach, leveraging a fine-tuned embedding model to generate embeddings for intentions and concepts. This fine-tuning ensured that intention embeddings were closer to their corresponding positive concept embeddings.}

As shown in Table~\ref{tab:conceptualization_prediction}, the results demonstrate that our conceptualization prediction method, based on the intention knowledge graph, achieves superior performance compared to LLMs regarding both ranking accuracy and inference time.

\begin{table}[]

\resizebox{\linewidth}{!}{
\begin{tabular}{@{}llllll@{}}
\toprule
 & MRR & Hit@1 & Hit@3 & Hit@10 & Inf. Time \\ \midrule
Llama3-8B & 0.3224 & 0.3023 & 0.3192 & 0.3449 & 2,069.96ms \\
Mistral-7B & 0.1110 & 0.0894 & 0.1001 & 0.1359 & 3,005.63ms \\
Flan-T5 & 0.0294 & 0.0058 & 0.0175 & 0.0564 & 1,790.91ms \\ \midrule
RIG (ours) & \textbf{0.4259} & \textbf{0.2476} & \textbf{0.5170} & \textbf{0.7906} & \textbf{181.82ms} \\ \bottomrule
\end{tabular}}
\vspace{-0.3cm}
\caption{The ranking performance and inference time on the task of conceptualization prediction.}
\vspace{-0.4cm}
\label{tab:conceptualization_prediction}
\end{table}

\paragraph{Product Recovery }
{\jx We constructed a session-intention pair dataset in previous sessions, including product IDs, descriptions, and user intentions. 
Here, we construct a new benchmarking dataset of the product-intention pairs from all session-intention pairs by assuming the items within the session share the same intentions as the sessions.
These product-intention pairs were randomly split into training, validation, and test sets with an 8:1:1 ratio. This dataset served as the basis for evaluating the ability of different methods to recover relevant intentions for given products.
To ensure a fair comparison, we focused on 1,203 overlapping products between the test set of the product-intention pairs in RIG and Folkscope. For these overlapped products, we compared the intentions generated by Folkscope and RIG. Rankings and evaluations were conducted on this same set of overlapping products, enabling a direct and balanced comparison of intention quality between the two systems. This methodology ensured the evaluation results reflected each method's capability to generate relevant and accurate intentions. Table~\ref{tab:product_recovery} provides the detailed evaluation results.
We used pre-computed intention and product embeddings as input to a Multi-Layer Perceptron (MLP) scoring model to train our model. The MLP was trained using Noise Contrastive Estimation (NCE) loss to distinguish between relevant and irrelevant intentions.
We assessed the model's performance on the test set during the evaluation phase by analyzing one positive sample against ten negative samples. Cosine similarity scores were computed for ranking intentions, enabling precise comparisons across methods.
Finally, we compared our method with Folkscope under identical experimental settings. The results, presented in Table~\ref{tab:product_recovery}, demonstrate that our knowledge graph significantly outperforms Folkscope in recovering relevant products from intentions, highlighting the superior efficacy of our approach.}

\begin{table}[t]
\resizebox{\linewidth}{!}{\begin{tabular}{@{}llllll@{}}
\toprule
& \# Intentions & (Ovlp.) & \# Products & (Ovlp.) \\
\midrule
FolkScope&1,846,715 &67,789 & 211,372 &1,203\\
RIG &295,620 &4,829 &453,124 &1,203\\
\midrule
\midrule
 & MRR & Hit@1 & Hit@3 & Hit@10  \\ \midrule
FolkScope(overlap)  &0.2808	 &0.0977	 &0.2816	 &0.9096
 \\ 
RIG (overlap) & \textbf{0.3161}	&\textbf{0.1257}	&\textbf{0.3453}	&\textbf{0.9263}\\ \midrule

FolkScope (full) & 0.2779	&0.0947	&0.2782	&0.9071 \\ 

RIG (full) & \textbf{0.3025}	&\textbf{0.1188}	&\textbf{0.3147}	&\textbf{0.9072}\\ \bottomrule
\end{tabular}
}
\vspace{-0.3cm}
\caption{The performance on product recovery. Ovlp. stands for overlap.}
\vspace{-0.4cm}
\label{tab:product_recovery}
\end{table}

\section{Extrinsic Evaluation}
 To demonstrate the practical utility of our relational intention knowledge graph, we evaluate its effectiveness in enhancing session-based recommendation, a key application domain for e-commerce platforms. Using the Amazon M2 dataset, we show how the rich semantic relationships captured in our knowledge graph can improve recommendation quality compared to state-of-the-art methods.

\begin{table*}[t]

\renewcommand\arraystretch{1.5}
\resizebox{\textwidth}{!}{
\begin{tabular}{l|l|cccccccccccccc|c}
\toprule[1.3pt]
Datasets  & Metric    & FPMC & GRU4Rec & BERT4Rec & SASRec & SASRecF & CORE & SR-GNN & GCE-GNN & DIF-SR & FEARec & DGNN & ISRec & Satori & KA-MemNN  & RIGRec \\ \midrule[1.3pt]
\multicolumn{1}{c|}{\multirow{6}{*}{M2 (UK)}} & Recall@5  & 0.2523 & 0.2792 & 0.1899 & 0.3075 & 0.2957 & 0.2990   & 0.2928   &  \underline{0.3130} & 0.3128  & 0.3088 & 0.3021 & 0.3073 & 0.2973 & 0.2932 & $\textbf{0.3342}^*$        \\
 & Recall@10 &0.3121 & 0.3469 & 0.2641 & 0.3964 & 0.3713 & 0.3949 &  0.3678 & \underline{0.4001} & 0.3990 & 0.3941 &  0.3882 & 0.3981 & 0.3821 & 0.3781 &    $\textbf{0.4229}^*$     \\
 & Recall@20 & 0.3696 & 0.4108 & 0.3349 & 0.4723 & 0.4406 & \underline{0.4768}  & 0.4381 & 0.4726 & 0.4739 & 0.4691 & 0.4563 & 0.4754 & 0.4436 & 0.4419 & $\textbf{0.5003}^*$    \\
 & Recall@50 & 0.4389 & 0.4865 & 0.4197 & 0.5621 & 0.5245  & \underline{0.5697} & 0.5171 & 0.5542 & 0.5598  & 0.5552 &  0.5584 & 0.5676 & 0.5348 & 0.5295 & $\textbf{0.5863}^*$    \\
 & Recall@100 & 0.4841 & 0.5346 & 0.4744 & 0.6159 & 0.5771  & \underline{0.6223} & 0.5662 & 0.6032 & 0.6171  & 0.6100 & 0.6072 & 0.6201 & 0.5931 & 0.5847 & $\textbf{0.6398}^*$    \\
 & NDCG@5    & 0.1933 & 0.2118 & 0.1260 & 0.2121 & 0.2208     & 0.1673 & 0.2195 & \underline{0.2214} & 0.2171 & 0.2138 &  0.2201 & 0.2207 & 0.2189 & 0.2163 &    \textbf{0.2214}    \\
 & NDCG@10   & 0.2126 & 0.2327 & 0.1501 & 0.2406 &  0.2432   & 0.1985 & 0.2418 & 0.2441 & \underline{0.2451} & 0.2415 &  0.2431 & 0.2438 & 0.2425 & 0.2386 & $\textbf{0.2503}^*$  \\
 & NDCG@20   & 0.2272 & 0.2499 & 0.1682 & 0.2598 &   0.2634  & 0.2193 & 0.2616 & 0.2626 & \underline{0.2641} & 0.2605 &  0.2637 & 0.2633 & 0.2623 & 0.2591 & $\textbf{0.2703}^*$  \\ 
 & NDCG@50   & 0.2411 & 0.2648 & 0.1848 & 0.2679 &   0.2801  & 0.2379 & 0.2784 & 0.2807 & \underline{0.2812} & 0.2777 &  0.2797 & 0.2795 & 0.2793 & 0.2737 & $\textbf{0.2877}^*$  \\ 
 & NDCG@100   & 0.2484 & 0.2718 & 0.1937 & 0.2862 &   0.2887  & 0.2472 & 0.2865 & 0.2871 & \underline{0.2891} & 0.2866 & 0.2867 & 0.2881 & 0.2864 & 0.2843 & $\textbf{0.2957}^*$  \\ 
 \bottomrule[1.3pt]
\end{tabular}}
\vspace{-0.3cm}
\caption{Performance comparison with baselines. The best and second-best results are shown in bold and underlined fonts. "$*$"
represents the significant improvement over the best baseline with p-value < 0.05.}
\vspace{-0.5cm}
\label{recommendation performance}
\end{table*}

\paragraph{Data Preparation}
We use the complete English subset of the Amazon M2 dataset integrated with our constructed knowledge graph. Following standard practice, we partition all sessions into training, validation, and test sets at an 8:1:1 ratio. To preserve the integrity of the dataset, we use the original preprocessing without filtering, thus avoiding potential session loss or artificial item connections.

The integration of a large-scale commonsense knowledge graph (containing over 351 million edges) with conventional recommendation systems presents significant challenges. To address this, we develop a meta-path approach to extract actionable item relationships from our knowledge graph. First, we identify session pairs connected through intention-level relationships (either conceptual or temporal). To ensure relationship quality, we retain only session pairs that either (1) share at least six distinct meta-paths through commonsense relation nodes or (2) demonstrate complete reachability from one session's concept nodes to the other's.
From these high-quality session pairs, we construct a weighted item graph $G=(V, E)$, where nodes represent products and edges represent their co-occurrence relationships within semantically connected sessions. Edge weights correspond to co-occurrence frequency, providing a measure of relationship strength. This approach effectively distills our comprehensive knowledge graph into a focused item relationship network tailored for recommendation tasks.

% Since all of the current session recommendation paradigms are difficult to integrate with a million-level commonsense knowledge graph, as we constructed in an end-to-end manner,
% we build upon an item relation graph based on it with meta-path methods to describe product relations. Specifically, we first gather all 1-hop session pairs where concepts or temporal relations directly connect their intentions. To alleviate redundant connections and noise, we only keep session pairs if (1) they have no less than six distinct meta paths passed through commonsense relation nodes or (2) one can reach the other from all of its 1-hop concept nodes. Then, we construct a weighted item graph $G=(V, E)$, where $V$ is the node-set, $E$ is the edge set generated by linked items in all sampled session pairs, and the co-occurrence frequency is as weight. Here, we focus on evaluating the effectiveness of relations derived from our session-intention KG, leaving the exploration of optimal sub-graph sampling strategy to be explored in future work.  

\paragraph{RIGRec: Intention-Enhanced Recommendation Model} 

We develop RIGRec, a novel recommendation model that leverages the rich semantic relationships captured in our intention knowledge graph. The model employs graph convolution to derive informative item representations:$\bm{E}^{l+1} = \bm{A}\bm{E}^{l},$
where $\bm{A}$ represents the adjacency matrix of graph $G$, and $\bm{E}\in \mathbb{R}^{N\times d}$ stores the $d$-dimensional embeddings for all $N$ items. We implement lightweight convolution operations and sum pooling to enhance computational efficiency. After $L$ convolution layers, the resulting representations $\bm{E}^{L}$ capture both item characteristics and their intention-based relationships.
For session modeling, we adopt SASRec's self-attention architecture, which effectively aggregates item representations within a session to model user preferences. This creates an end-to-end recommendation system that seamlessly integrates intention-level reasoning with sequential pattern recognition, enabling more contextually appropriate recommendations.

% Given the relation $G$, one can obtain item representations with various graph representation methods. We adopt the simple yet effective graph convolution operation to learn informative item representations, which is formulated as 
%     $\bm{E}^{l+1} = \bm{A}\bm{E}^{l},$
% where $\bm{A}$ is the adjacency matrix of $G$, $\bm{E}\in \mathbb{R}^{N\times d}$ is the $d$-dimensional embedding dictionary of all items. Canonical methods are utilized during graph representation learning, including light-weight convolution and sum pooling. After $L$-th convolution, the knowledge graph-based representations $\bm{E}^{L}$ can be directly utilized in the recommendation tasks for enhancing performance. 
% To achieve end-to-end recommendation, we designed a Relational Intention Knowledge Graph-based recommender named RIGRec, which seamlessly integrates the graph representations learning module into the session recommendation framework. Concretely, the widely used attention-based method, SASRec, is adopted as our session encoder, which aggregates the learned knowledge graph-based item representations within each session for user preference estimation.

\begin{figure*}[t]
    \centering
    \subfigure{
    \includegraphics[width=0.43\textwidth]{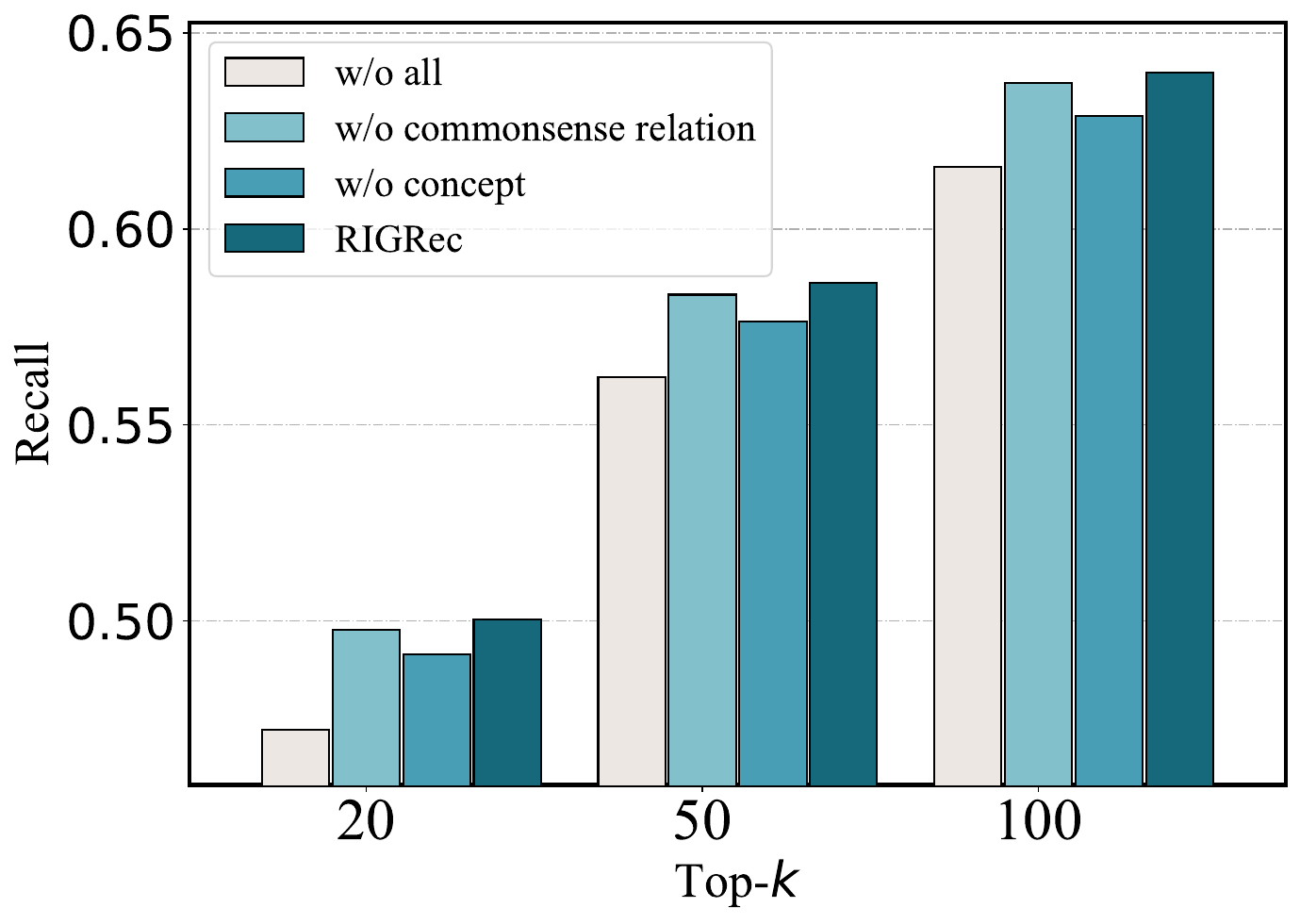}}
    \subfigure{
    \includegraphics[width=0.43\textwidth]{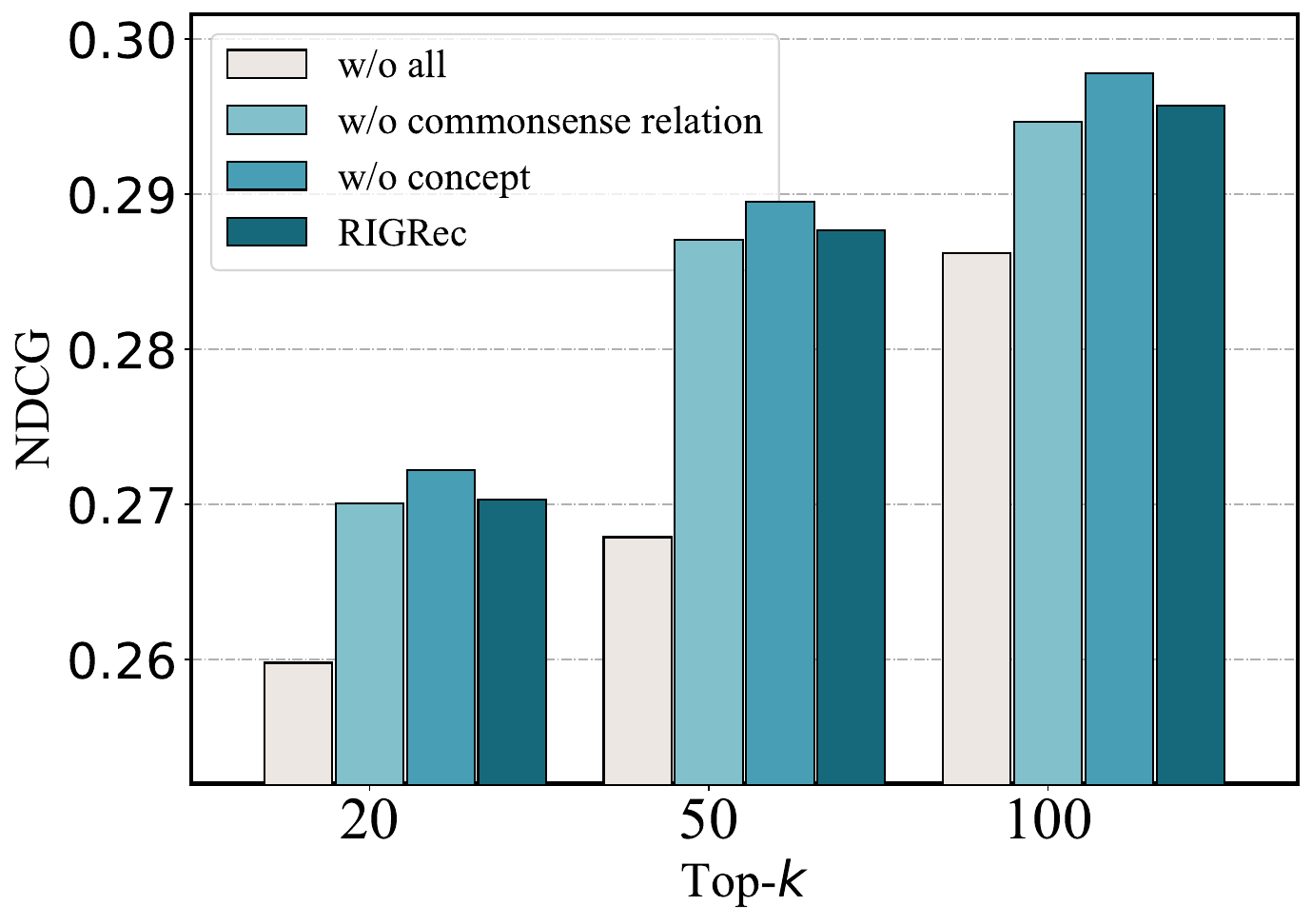}}
    \vspace{-0.7cm}
    \caption{Ablation results of different variants. This demonstrates that our intention knowledge graph significantly enhances recommendation performance compared to SASRec. Both intention conceptualization and concept relations effectively improve results, with each type of relation contributing uniquely to different metrics. This highlights the importance of incorporating diverse nodes and relations in the knowledge graph.}
    \label{ablation}
    \vspace{-0.3cm}
\end{figure*}

\paragraph{Baselines and Evaluation Metrics} 
We compare our model with following ten representative and state-of-the-art methods, covering (1) the classical method FPMC \cite{fpmc}, (2) the RNN-based method GRU4Rec \cite{gru4rec}, (3) the predominant attention-based methods including BERT4Rec \cite{bert4rec}, SASRec \cite{sasrec}, CORE \cite{core} and FEARec \cite{fearec}, (4) graph-based methods including SR-GNN \cite{srgnn} and GCE-GNN \cite{gcegnn}, (5) side information fusion methods including SASRecF \cite{sasrec} and DIF-SR \cite{dif-sr}. We exclude some state-of-the-art methods like FAPAT \cite{FAPAT} due to the need for massive support resources or exponential computation complexity. {\jx Additionally, we compare our approach with  state-of-the-art intention-aware recommendation models: DGNN \cite{li2023exploiting}, ISRec \cite{li2022intention}, Satori \cite{chen2022sequential}, and KA-MemNN \cite{zhu2021learning}.}

We employ two standard metrics in the field of recommender systems, including Recall at a cutoff top $k$ (Recall$@k$) and Normalized Discounted Cumulative Gain at a cutoff top $k$ (NDCG$@k$). 
We rank the ground-truth item alongside all candidates to ensure an unbiased evaluation rather than adopting the negative sampling strategy. We report the averaged metrics over 5 runs with the commonly utilized $k \in \{5,10,20,50,100\}$. To ensure unbiased evaluation, we rank the ground-truth item alongside all candidate items rather than using negative sampling. Results represent averages across five independent runs, with statistical significance determined through unpaired t-tests (p < 0.05).
The implementation details are in the Appendix \ref{sec:sess_impl_details}.

% \begin{figure*}[t]
%     \centering
%     \subfigure[]{
%     \includegraphics[width=0.45\textwidth]{m2_recall.pdf}}
%     \subfigure[]{
%     \includegraphics[width=0.45\textwidth]{m2_ndcg.pdf}}
%     \vspace{-0.5cm}
%     \caption{Ablation results of different variants. This demonstrates that our intention knowledge graph significantly enhances recommendation performance compared to SASRec. Both intention conceptualization and concept relations effectively improve results, with each type of relation contributing uniquely to different metrics. This highlights the importance of incorporating diverse nodes and relations in the knowledge graph.}
%     \label{ablation}
% \end{figure*}

\paragraph{Performance Comparison}

As shown in Table \ref{recommendation performance}, RIGRec consistently outperforms all baseline methods across nearly all evaluation metrics, with statistically significant improvements over the strongest competitors. These results demonstrate the substantial benefits of incorporating intention-level relationships.

The performance gap between RIGRec and SASRec—which serves as RIGRec's backbone architecture—is particularly notable. This difference directly quantifies the value added by our intention knowledge graph, confirming that modeling user intentions and their relationships provides critical information beyond what can be captured by sequential patterns alone.

Graph-based methods like SR-GNN and GCE-GNN, while theoretically capable of modeling complex item relationships, show inferior performance compared to our approach. This stems from their focus on direct item transitions without consideration of the underlying user intentions driving these transitions. By explicitly modeling intention-level relationships, RIGRec captures deeper semantic connections between items, resulting in more contextually appropriate recommendations.

The mixed performance of side information fusion methods (DIF-SR and SASRecF) highlights the challenges of effectively integrating auxiliary information into recommendation systems. While DIF-SR achieves competitive results through careful information fusion, SASRecF shows performance degradation compared to the base SASRec model in terms of Recall@k. This aligns with findings that inappropriate information integration strategies can degrade model effectiveness. Our approach, by contrast, leverages large language models to extract high-quality intention information, filtering out noise and irrelevant features that might otherwise compromise recommendation quality.

\paragraph{Ablation Study}
To isolate the contributions of different components in our knowledge graph, we conduct ablation studies comparing RIGRec against three variants:
(1)\textbf{ w/o all}: Removes the entire item graph derived from our knowledge graph (equivalent to standard SASRec).
(2) \textbf{w/o concept}: Removes edges derived from concept-mediated paths.
(3)\textbf{ w/o commonsense relation}: Removes edges derived from commonsense relation paths.

The ablation study reveals three insights: (1) The intention knowledge graph substantially improves recommendation quality beyond what sequential patterns alone can achieve; (2) Both conceptualization and commonsense relations provide complementary signals that enhance performance; and (3) These relation types contribute differently to evaluation metrics—conceptualization improves recall by identifying broadly relevant items, while commonsense relations enhance NDCG by capturing fine-grained semantic connections for better ranking. These findings confirm that modeling diverse intention relationships leads to more contextually appropriate recommendations by providing a more comprehensive understanding of user behavior.

% Figure \ref{ablation} presents the results of the ablation studies; we interpret the results with the following discoveries. First, our constructed intention knowledge graph is of vital use to produce helpful item representations for downstream recommendation tasks. As we can see, the performance of our model exceeds that of SASRec by a large margin. Second, both types of information contained in intention conceptualization and concepts relation prove efficient. It can be confirmed that the two variants solely with corresponding edges can outperform w/o all variants.
% Third, distinct relations provide dissimilar contributions to each metric, w/o commonsense relation presenting high Recall metrics. At the same time, w/o conceptualization performs better in NDCG metrics, which reaffirms the significance of constructing multi-type nodes and relations in our intention knowledge graph. 

\section{Conclusion}
% We present IGC-RC, a framework for automatically constructing intentional commonsense knowledge graphs from user behaviors. Using this framework, we built the Relational Intention Knowledge Graph (RIG) and validated its quality through extensive evaluations. Results show that RIG significantly enhances the performance of state-of-the-art session recommendation models.

We introduce IGC-RC, a framework for automatically constructing knowledge graphs that model relationships between user intentions in e-commerce contexts. Our Relational Intention Graph (RIG) captures temporal, causal, and conceptual connections between user goals, demonstrating exceptional quality in both intrinsic evaluations and recommendation tasks. By bridging the gap between observable behaviors and underlying intentions, RIG enables more accurate prediction of user needs and significantly enhances recommendation performance. This advancement represents an important step toward more human-like understanding in intelligent e-commerce systems.

\section*{Limitations}
The intention generation process relies on GPT-3.5, which may introduce additional computational overhead. Future work could explore more efficient language models to streamline this component.
Our framework is evaluated using the Amazon M2 dataset, which is specific to e-commerce. The applicability of the proposed method to other domains remains to be tested. The current implementation focuses on the English subset of the dataset. Extending the framework to support multiple languages could enhance its versatility. While we incorporate commonsense relations such as temporality and causality, the scope of relation types is limited. Incorporating a broader range of relational categories may improve the knowledge graph's comprehensiveness.
{ We would like to provide more validation if more suitable datasets are available. However, most public datasets are desensitized and anonymized, making generating intention based on the anonymized ID features hard. 
Besides, our utilized M2 dataset is a mixed-type dataset, which already contains multi-typed items (sports, beauty, baby, etc.). 
The item and session sizes also exceed general research works. }

\section*{Ethics Statement}
This study ensures the responsible use of data and technology by utilizing the anonymized Amazon M2 dataset, which safeguards user privacy and complies with data protection regulations. We have implemented measures to prevent the inclusion of any personally identifiable information (PII). Additionally, we acknowledge potential biases in the dataset and have taken steps to mitigate them through standard preprocessing techniques. Our use of large language models focuses on enhancing user experience without manipulating behavior, and we advocate for transparency in deploying intention knowledge graphs within e-commerce platforms.

\bibliography{acl_latex}

\newpage
\appendix

\section{Further Details on Experiments}
\subsection{Human Annotation}
{
We collected annotations for 3,000 session-intention pairs, each evaluated by three annotators. The inter-annotator agreement scores were 0.91 for plausibility and 0.74 for typicality. 
Three annotators independently evaluated each of the 1,000 intention-intention discourse pairs for intention relation classification, achieving an overall inter-annotator agreement of 0.69. All raw annotation data will be made publicly available to support future research in this area. 
We use exactly the same annotation guidelines and criteria from the FolkScope paper, which can ensure the same standard of annotations on plausibility and typicality.  
}
\subsection{Concept Prediction}
{ 
We conducted experiments to compare different methods for predicting conceptualized intentions, where the goal is to predict corresponding concepts for a given purpose. 

The first method employed a generative approach using LLMs, specifically Mistral-7B-Instruct-v0.3, Meta-Llama-3-8B-Instruct, and flan-t5-xl. This approach generated concepts in the same way as our RIG construction process. During testing, we created a candidate pool containing both true and false concepts for each intention. The LLMs generated ten concepts per intention, and matching concepts were ranked first while maintaining their generation order. We used negative sampling to create a candidate pool of 500 concepts. Importantly, we did not fine-tune the LLMs to maintain consistency with the approach used in RIG's construction. 

Our proposed approach's second method utilized an embedding-based model (bge-base-en-v1.5) to transform intentions and concepts into embeddings and compute their cosine similarities. We fine-tuned the embedding model using contrastive learning, incorporating cross-entropy loss to improve matching performance. During testing, we created a candidate pool of 500 concepts and ranked them based on their cosine similarities with the given intention. 

We constructed a dataset of intention-concept pairs from RIG for our experimental setup. We split it into training, validation, and testing sets with an 8:1:1 ratio, resulting in 147,801 intention-concept pairs in the test set. We conducted our experiments using an Nvidia RTX A6000 GPU. The models were evaluated using standard metrics, including MRR, Hit@1, Hit@3, Hit@10, and inference speed. The results demonstrated that our embedding-based method achieved superior prediction accuracy and computational efficiency performance compared to the LLM approach.}

\subsection{Intention Prediction}

{
The input of this task is a session, and the output of this task is the ranking of the ground-truth intention over a pool of negative intentions. The goal of this task is to rank the correct intention higher than the incorrect intentions. Because the input of this task is a session, so we use the SASRec as the backbone model. We used the all-MiniLM-L6-v2 model to generate embeddings for session items and intentions. These item embeddings initialized the item embedding matrix in SASRec. During training, the parameters of session encoders and item embeddings are tuned. TripletMarginLoss minimized the distance between session and positive intention embeddings while maximizing the distance to negative intentions. Random sampling of positive and negative samples, followed by backpropagation and early stopping, was used to optimize the model. }

\subsection{Product Recovery Benchmark}
{
We use the same evaluation method to ensure the fairness of the evaluation.  
Table \ref{tab:product_recovery}  shows the evaluation based on 1,203 overlapping products between the two graphs. For these identical products, we compared the intentions generated by Folkscope and RIG, respectively. The rankings and evaluations were conducted using this same set of overlapping products, allowing for a direct comparison of intention quality between the two systems. This methodology ensures a fair and balanced assessment of each method’s ability to generate relevant intentions.    

Using the same sentence embedding model (BGE), we acquired pre-computed intention and product embeddings. Then, we used the intention embeddings as input and the product embeddings as output to train a Multi-Layer Perceptron (MLP) scoring model using Noise Contrastive Estimation (NCE) loss. In the evaluation phase, for each intention, we analyzed one positive sample against 10 negative samples by calculating the cosine similarity scores of their embeddings to the target embedding and subsequent rankings, where the rank was determined by the count of negative samples scoring higher than the positive sample plus one. 

We have identified and addressed key limitations in prior work on intention generation using large language models. \cite{DBLP:journals/corr/abs-2402-14901} highlighted two significant issues with FolkScope's intentions: property-ambiguity and category-rigidity. These issues primarily stem from two factors. First, using a relatively weak language model (OPT-30B) for intention generation limited its ability to produce high-quality outputs. Second, the reliance on ConceptNet relations for prompt construction introduced constraints, as some relations (e.g., "made of") were not well-suited for generating diverse and meaningful user intentions.

Our work tackles these limitations through two key improvements. First, we employ a more capable language model, Llama3-8B-instruction, to enhance the quality of intention generation. Second, we remove the reliance on ConceptNet relations in prompts. Instead, we leverage the advanced capabilities of modern language models to capture open-ended intentions, enabling the generation of more natural and diverse user intentions without being restricted by predefined relation types.

We have achieved improved intention quality by addressing these issues from the outset. 
To validate these improvements, we followed the evaluation methodology outlined in \cite{DBLP:journals/corr/abs-2402-14901}, using product recovery benchmarks. As shown in Table \ref{tab:product_recovery}, our approach demonstrates superior performance compared to the baseline model.
}
\subsection{Implementation Details of Session Recommendation}
\label{sec:sess_impl_details}
{ For a fair comparison, the dimension of item embedding is set to 64 for all methods. Grid search strategy is applied to determine the optimal configuration of standard parameters, involving the learning rate in $\{1e^{-2},1e^{-3},1e^{-4}\}$, the dropout rate in $\{0,0.1,0.2,0.3,0.4\}$, the loss function in $\{$BPR loss, Binary Cross Entropy loss, Cross Entropy loss$\}$ and the coefficient of $L2$ regularization in $\{0,1e^{-2},1e^{-3},1e^{-4}\}$.}

\begin{table}[t]
\centering

\begin{tabular}{@{}l|c|cc@{}}
\toprule
N-gram & FolkScope & RIG & w/ConceptNet Rel \\ \midrule
2-gram & 0.0307 & \textbf{0.5274} & {\jx 0.3759}\\
3-gram & 0.0480 & \textbf{0.7931} & {\jx 0.5806}\\
4-gram & 0.0648 & \textbf{0.9100} &  {\jx 0.6997}\\
5-gram & 0.0837 & \textbf{0.9623} & {\jx 0.7751} \\
6-gram & 0.1046 & \textbf{0.9833} & {\jx 0.8264} \\ \bottomrule
\end{tabular}
\caption{N-gram diversity scores of intentions extracted from FolkScope and RIG. The results validate that RIG generates more diverse intentions by removing ConceptNet relation constraints.}
\label{tab:ngram_diversity}
\end{table}

\section{Large Language Model Generation Prompts}

The following Figure \ref{fig:generation_prompts} and Figure \ref{fig:conceptualization_prompts} denote the prompts that we used to generate the intentions and concepts. 

\begin{figure}[t]
\small
\begin{AIbox}{{Session Intention Generation}}
Below is a user's chronological record list: [SESSION]\\
Explain the basic intentions of this user exactly. Output several different intentions one by one to answer the following question: Users buy these items because they want to:\\
intention 1: \{a simple verb phrase within 10 words\}\\
intention 2: \{a simple verb phrase within 10 words\}
...
\end{AIbox}
\caption{This figure shows the prompts we use to make LLM understand and generate intentions from user sessions.}
    \label{fig:generation_prompts}
\end{figure}

% \begin{figure}[h]
%     \centering
%     \includegraphics[width=\linewidth]{figures/generation_prompts.pdf}
%     \caption{This figure shows the prompts we use to make LLM understand and generate intentions from user sessions.}
%     \label{fig:generation_prompts}
% \end{figure}
\begin{figure}[ht]
    \small
   \begin{AIbox}{{Abstract Intention Generator}}
I will give you an INTENTION. You need to give several phrases containing 1-3 words for the ABSTRACT INTENTION of this INTENTION. You must return your answer in the following format: phrases1,phrases2,phrases3,...., which means you can't return anything other than answers.
These abstract intention words should fulfill the following requirements:\\
1. The ABSTRACT INTENTION phrases can well represent the INTENTION.\\
2. The ABSTRACT INTENTION phrases don't have a lot of less relevant word meanings. For example, ``spring'' is not a good abstract intention word because it can represent both a coiled metal device and the season of the year.\\
3. The ABSTRACT INTENTION phrases of the same INTENTION cannot be semantically similar to each other. For example, health and wellness are two close synonyms, so they can't be together.\\
INTENTION: Moisturize dry skin while enjoying a special effect bath.\\
Your answer: hydration, skincare\\
INTENTION: Create a festive atmosphere for a Christmas party.\\
Your answer: party planning, celebration, decorations, holiday spirit\\
INTENTION: [INTENTION].\\
Your answer:
\end{AIbox}
    \caption{This figure shows our prompts to make LLM conceptualize the user intentions.}
    \label{fig:conceptualization_prompts}
\end{figure}

% \begin{figure}[h]
%     \centering
%     \includegraphics[width=\linewidth]{figures/conceptualization_prompts.pdf}
%     \caption{This figure shows our prompts to make LLM conceptualize the user intentions.}
%     \label{fig:conceptualization_prompts}
% \end{figure}

\begin{table*}[p]

\resizebox{\textwidth}{!}{\begin{tabular}{@{}p{4cm}p{4cm}p{9cm}@{}}
\toprule
Intention 1 & Intention 2 & Assertion \\ \midrule
Relieve discomfort and soothe itching caused by haemorrhoids. & Purchase unscented baby wipes for sensitive skin. &\thinspace{\color{teal} People} relieve discomfort and soothe itching caused by haemorrhoids, {\color{teal}and simultaneously, they} purchase unscented baby wipes for sensitive skin. \\\midrule
Make coffee at home. & Enjoy a variety of coffee flavors at home. & \thinspace{\color{teal}People} make coffee at home {\color{teal}usually after they} enjoy a variety of coffee flavors at home.\\\midrule
Dress up as Lara Croft for a costume party or event. & Have fun with Halloween-themed party games. & \thinspace{\color{teal}People} dress up as Lara Croft for a costume party or event, {\color{teal}and simultaneously, they} have fun with Halloween-themed party games. \\\midrule
Find a cream that provides fast and numbing relief from haemorrhoid symptoms. & Use advanced moisture absorption technology. & \thinspace{\color{teal}People} find a cream that provides fast and numbing relief from haemorrhoid symptoms {\color{teal}because they} use advanced moisture absorption technology. \\\midrule
Maintain personal hygiene and cleanliness. & Purchase a razor handle and blade refills for men's shaving. &\thinspace {\color{teal}People} maintain personal hygiene and cleanliness{\color{teal}, as a result, they} purchase a razor handle and blade refills for men's shaving.  \\ \bottomrule
\end{tabular}}
\caption{This table presents two candidate intentions and related assertions. The assertions provide an interpretive summary of the relationship between the paired intentions. The templates mapping from triples to assertions are marked in green.}
\label{tab:vera_examples}
\end{table*}

\section{Annotation Questions}

Here, we give three examples of annotation questions from our questionnaire for the Amazon Mechanical Turk. 
They are for the intentions quality annotation, intention relation classification annotation, and intention conceptualization classification. 
\begin{table*}[th]
\centering

\resizebox{\linewidth}{!}{
\begin{tabular}{@{}lcccccccccc@{}}
\toprule
Model & R@5 & R@10 & R@20 & R@50 & R@100 & N@5 & N@10 & N@20 & N@50 & N@100 \\ \midrule
GRU4Rec & 0.2792 & 0.3469 & 0.4108 & 0.4865 & 0.5346 & 0.2118 & 0.2327 & 0.2499 & 0.2648 & 0.2718 \\
RIG-GRU4Rec & 0.2923 & 0.3852 & 0.4665 & 0.5562 & 0.6028 & 0.2126 & 0.2367 & 0.2474 & 0.2693 & 0.2746 \\
\midrule
SASRec & 0.3075 & 0.3964 & 0.4723 & 0.5621 & 0.6159 & 0.2121 & 0.2406 & 0.2598 & 0.2679 & 0.2862 \\
RIGRec & \textbf{0.3342} & \textbf{0.4229} & \textbf{0.5003} & \textbf{0.5863} & \textbf{0.6398} & \textbf{0.2214} & \textbf{0.2503} & \textbf{0.2703} & \textbf{0.2877} & \textbf{0.2957} \\
\bottomrule
\end{tabular}}
\caption{Performance comparison of different backbone architectures with and without our knowledge graph.}
\label{tab:backbone_comparison}
\end{table*}

\section{Evaluating Diversity of Intention}

As shown in Figure \ref{tab:ngram_diversity}, we further compare the diversity of the generated intentions by using the n-gram diversity. It is defined as the ratio of the  unique n-gram counts to all n-gram counts:

\begin{equation}
\text{Diversity}(D, n) = \frac{\# \text{ unique } n\text{-grams in } D^{\oplus}}{\# \text{ n-grams in } D^{\oplus}}
\end{equation}
Where $D^{\oplus}$ denotes the dataset $D$ concatenated into a single string. We use six as the 
maximum n-gram length. This method captures repeated sequences in addition to single-token diversity. 

We measured the diversity of the corpus formed by the intentions extracted from ForkScope and RIG. {\jx For the intention generation with conceptnet relation, we randomly sample 100 intentions derived from each of 17 conceptnet relations, totaling 1700 intentions. For RIG and Folkscope, we sample 1700 intentions from each of them to compute the n-gram diversity.} The results demonstrate that RIG's N-gram diversity of intentions is significantly higher than ForkScope's. These findings validate our claim that removing ConceptNet relation constraints {\jx and using a better generation model} generates more diverse intentions.

\section{Analysis of Different Backbones for RIGRec}
{\jx
The choice of backbone architecture is an important consideration when implementing our approach. In our main experiments, we primarily utilized SASRec as the backbone for RIGRec due to its simplicity, robustness, and widespread adoption as a baseline in session-based recommendation research. This choice provides clear interpretability of the improvements brought by our intention knowledge graph.

However, to demonstrate the generalizability of our approach, we conducted additional experiments using GRU4Rec as an alternative backbone. Table~\ref{tab:backbone_comparison} presents the performance comparison between the original backbones (GRU4Rec and SASRec) and their RIG-enhanced versions (RIG-GRU4Rec and RIGRec).

The results demonstrate that our approach consistently improves performance regardless of the backbone architecture. When applied to GRU4Rec, our intention knowledge graph enhances performance across all metrics, with particularly notable improvements in Recall@20 (13.6\% improvement) and Recall@50 (14.3\% improvement). This confirms that the benefits of our relational intention knowledge graph are not limited to a specific recommendation architecture.

Nevertheless, the RIGRec model (based on SASRec) still achieves the best overall performance. This can be attributed to SASRec's inherent advantages in capturing long-range dependencies through its self-attention mechanism, which may better complement the high-level intention information provided by our knowledge graph. The results suggest that while our approach generalizes well to different backbones, the choice of backbone can still influence the absolute performance levels achieved.}

\section{Case Studies}{\jx

To illustrate the practical effectiveness of our RIG approach, we present three representative case studies in Table~\ref{tab:case_study}. These examples highlight how our model leverages intention relationships to generate more contextually relevant recommendations compared to baseline methods.

\begin{table*}[t]
\centering

\resizebox{\textwidth}{!}{%
\begin{tabular}{@{}p{5.5cm}p{5cm}p{6cm}p{6cm}@{}}
\toprule
\textbf{Session History} & \textbf{Generated Intentions} & \textbf{Baseline Recommendations (SASRec)} & \textbf{Our Model Recommendations (RIGRec)} \\ \midrule

Wireless Gaming Mouse; RGB Gaming Keyboard; Gaming Headset with Microphone 
& 
Set up a complete gaming peripherals system; Enhance gaming experience with RGB accessories; Improve communication during multiplayer games
& 
Ergonomic Gaming Chair; Gaming Mouse Pad; USB Hub; Mechanical Gaming Keyboard; RGB Gaming Mouse
& 
\textbf{Gaming Mouse Pad with RGB}; \textbf{Gaming Desk with Cable Management}; \textbf{Dual Monitor Stand}; \textbf{RGB LED Strip for Desk}; \textbf{Gaming Controller}
\\
\midrule

Baby Diapers Size 1 (240 Count); Baby Wipes Unscented; Diaper Rash Cream; Baby Bottle Sterilizer
& 
Prepare essential baby care items for a newborn; Maintain baby hygiene; Prevent and treat diaper rash; Ensure safe feeding equipment
& 
Baby Powder; Baby Wipes Sensitive; Baby Lotion; Baby Shampoo; Diaper Bag
& 
\textbf{Baby Bottles Anti-Colic}; \textbf{Bottle Drying Rack}; \textbf{Changing Pad}; \textbf{Baby Bottle Brush Set}; \textbf{Bottle Warmer}
\\
\midrule

Pumpkin Carving Kit; Halloween String Lights; Artificial Spider Web; Halloween Doorbell with Spooky Sounds
& 
Decorate home for Halloween; Create a spooky atmosphere; Prepare for trick-or-treaters; Host a Halloween party
& 
Halloween Window Decorations; Halloween Candy Bowl; Halloween Mask; Halloween Costume; Fog Machine
& 
\textbf{Halloween Door Wreath}; \textbf{Motion-Activated Halloween Props}; \textbf{Outdoor Halloween Projector Lights}; \textbf{Halloween Themed Serving Tray}; \textbf{Halloween Party Snack Bowls}
\\
\bottomrule
\end{tabular}}
\caption{Case Study: Example Sessions and Recommendations}
\label{tab:case_study}
\end{table*}

\subsection{Gaming Setup Completion}
In the first case, the user has purchased basic gaming peripherals (mouse, keyboard, and headset). The baseline model recommends similar gaming peripherals and accessories without considering the user's broader intention. In contrast, our RIGRec model identifies the user's goal of setting up a complete gaming system and recommends complementary products that enhance the overall gaming environment, such as a desk with cable management and RGB lighting solutions. This demonstrates how our model captures the conceptual relationships between products through the user's intentions.

\subsection{Baby Care Essentials}
The second case shows a user purchasing basic newborn care items. While the baseline recommends additional hygiene products that are similar to those already in the cart, our model recognizes the broader intention of preparing for a newborn's feeding needs. By identifying the connection between diaper changing and feeding essentials as part of comprehensive baby care, RIGRec suggests complementary products like anti-colic bottles and a bottle warmer that the baseline misses entirely.

\subsection{Halloween Preparation}
In the third case, the user is collecting Halloween decorations. The baseline model focuses on recommending additional decorative items and costumes. Our model, however, identifies the potential intention of hosting a Halloween party and recommends party-specific items like themed serving trays and snack bowls. This demonstrates how RIGRec's understanding of intention relationships (decorating for Halloween → hosting a Halloween party) enables it to anticipate future user needs that aren't explicitly indicated in the current session.

\subsection{Analysis of Intention Relations}
Table~\ref{tab:intention_relations} shows examples of how our model captures key intention relationships that help generate better recommendations. These intention pairs demonstrate the temporal and causal connections our knowledge graph identifies.

\begin{table*}[t]
\centering

\resizebox{\linewidth}{!}{%
\begin{tabular}{@{}lll@{}}
\toprule
\textbf{Initial Intention} & \textbf{Related Intention} & \textbf{Relation Type} \\ \midrule
Set up a gaming peripherals system & Create an immersive gaming environment & Causality \\
Decorate home for Halloween & Host a Halloween party & Temporal (Before→After) \\
Ensure safe feeding equipment & Prepare bottles for infant feeding & Synchronous \\
Upgrade PC components & Enhance gaming performance & Causality \\
Purchase cooking utensils & Prepare homemade meals & Temporal (Before→After) \\
\bottomrule
\end{tabular}}
\caption{Examples of Captured Intention Relations in RIG}
\label{tab:intention_relations}
\end{table*}

These cases illustrate how our intention knowledge graph helps bridge the gap between observed behaviors and underlying user goals, resulting in recommendations that better address the user's complete needs rather than simply suggesting similar products. The ability to identify conceptual relationships between different intentions enables our model to make more contextually appropriate and diverse recommendations.
}

\begin{figure*}[ht]
    \centering
    \includegraphics[width=\linewidth]{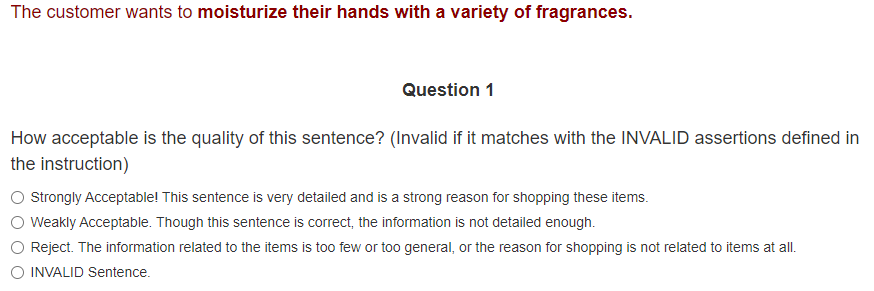}
    \caption{This figure shows an example annotation question for the quality of session intention generation.}
    \label{fig:intention_annotation}
\end{figure*}

\begin{figure*}[ht]
    \centering
    \includegraphics[width=\linewidth]{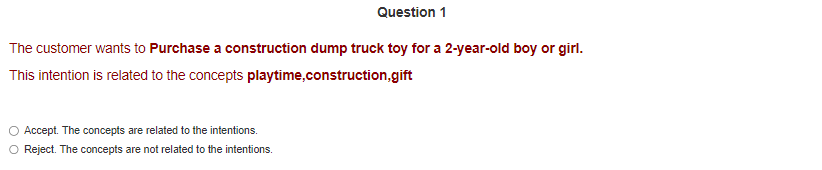}
    \caption{This figure shows an example annotation question for the quality of session intention conceptualization.}
    \label{fig:conceptualization_annotationi}
\end{figure*}

\begin{figure*}[ht]
    \centering
    \includegraphics[width=\linewidth]{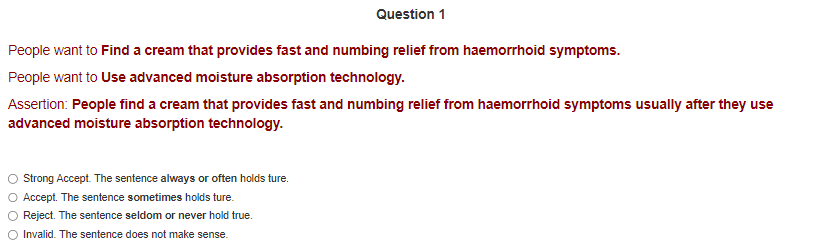}
    \caption{This figure shows an example annotation question for the quality of session intention relation classification.}
    \label{fig:intention classification}
\end{figure*}

\section{Extended Ablation Study on LLM Backbones for Intention Generation}
\label{sec:appendix_llm_ablation}

To assess the robustness of our framework and address concerns regarding the reliance on proprietary models (specifically GPT-3.5), we conducted an ablation study comparing different Large Language Models (LLMs) for the intention generation phase. We compared the current backbone (GPT-3.5) against two state-of-the-art open-weight models: \textbf{Llama3-8B-Instruct} and \textbf{Mistral-7B-Instruct}.

We evaluated the generated intentions based on \textbf{Plausibility} (whether the intention makes sense in the context) and \textbf{Typicality} (how common the intention is), alongside the inference cost.

\begin{table*}[h]
    \centering
    \begin{tabular}{lccc}
        \toprule
        \textbf{Model} & \textbf{Plausibility} & \textbf{Typicality} & \textbf{Cost (per 1K tokens)} \\
        \midrule
        \textbf{GPT-3.5 (Ours)} & \textbf{0.955} & \textbf{0.667} & $\sim$\$0.0015 \\
        Llama3-8B & 0.948 & 0.652 & Free (Self-hosted) \\
        Mistral-7B & 0.931 & 0.638 & Free (Self-hosted) \\
        \bottomrule
    \end{tabular}
    \caption{Performance and Cost Comparison of LLM Backbones. Our framework maintains high performance even with open-weight models.}
    \label{tab:llm_comparison}
\end{table*}

\paragraph{Analysis.} As shown in Table~\ref{tab:llm_comparison}, while GPT-3.5 achieves the highest performance, the gap between it and open-weight models is marginal. Llama3-8B trails by only 0.7\% in plausibility and 1.5\% in typicality. This demonstrates that our \textbf{IGC-RC} (Intention Generation, Conceptualization, and Relation Classification) framework is model-agnostic and can be effectively deployed using open-source models in resource-constrained environments or where data privacy requires local processing.

\section{Sensitivity Analysis of Vera Score Thresholds}
\label{sec:appendix_thresholds}

In the construction of the Relational Intention Graph (RIG), we utilized a Vera score threshold to filter low-quality edges. To validate our choice of threshold and analyze the trade-off between graph scale and edge acceptance quality, we performed a sensitivity analysis using thresholds of 0.85, 0.90 (our selected value), and 0.95.

\begin{table*}[h]
    \centering
    
    \begin{tabular}{lcc}
        \toprule
        \textbf{Threshold} & \textbf{Acceptance Rate} & \textbf{Number of Edges} \\
        \midrule
        0.85 & 79.8\% & 412.0 M \\
        \textbf{0.90 (Selected)} & \textbf{81.2\%} & \textbf{341.6 M} \\
        0.95 & 82.4\% & 218.0 M \\
        \bottomrule
    \end{tabular}
    \caption{Impact of Vera Score Thresholds on Graph Scale and Quality. The selected threshold (0.90) offers the optimal trade-off.}
    \label{tab:threshold_analysis}
\end{table*}

\paragraph{Analysis.} The results in Table~\ref{tab:threshold_analysis} highlight the rationale behind our selection. A threshold of 0.95 significantly reduces the graph density (dropping nearly 123 million edges) for a marginal gain in acceptance rate (1.2\%). Conversely, lowering the threshold to 0.85 introduces more noise. The 0.90 threshold offers the optimal balance, maintaining a high acceptance rate ($>80\%$) while preserving the massive scale necessary for training robust recommendation models.

It is worth noting that while manually curated graphs like Atomic2020 achieve higher acceptance rates ($\sim$86.8\%), they are limited in scale ($\sim$0.6M edges). RIG achieves comparable quality (within $\sim$5\%) while being over \textbf{500$\times$ larger} and fully automated.

\section{Inference Efficiency and Practicality Analysis}
\label{sec:appendix_efficiency}

A key advantage of our pre-computed Knowledge Graph approach (RIG) over direct LLM inference for recommendations is speed. We compared the inference latency of our RIG-enhanced recommendation model against a direct LLM-based recommender (using Mistral-7B).

\begin{table*}[h]
    \centering
    \begin{tabular}{lcccc}
        \toprule
        \textbf{Method} & \textbf{MRR} & \textbf{Hit@10} & \textbf{Inference Time} & \textbf{Speedup} \\
        \midrule
        Mistral-7B (Direct) & 0.5377 & 0.6763 & 3,625.63 ms & 1$\times$ \\
        \textbf{RIG (Ours)} & \textbf{0.5544} & \textbf{0.7260} & \textbf{3.01 ms} & \textbf{$\sim$1,200$\times$} \\
        \bottomrule
    \end{tabular}
    \caption{Efficiency Comparison (Inference Time vs. Performance). RIG achieves comparable accuracy with drastically lower latency.}
    \label{tab:efficiency_comparison}
\end{table*}

\end{document}